\documentclass[10pt,conference,twocolumn]{IEEEtran}
\IEEEoverridecommandlockouts

\usepackage{booktabs}
\usepackage{multirow}
\usepackage[colorlinks]{hyperref}

\usepackage{cite}
\usepackage{amsmath,amssymb,amsfonts}
\usepackage{algorithmic}
\usepackage{graphicx}
\usepackage{textcomp}
\usepackage{xcolor}
\usepackage{subfiles}
\usepackage{multirow}
\usepackage{tabularx}
\usepackage{listings}
\usepackage{multirow}
\usepackage{lipsum}
\usepackage{multicol}
\usepackage{float}
\usepackage{gensymb}
\usepackage{url}
\usepackage[justification=centering]{caption}

\makeatletter 
\newcommand{\linebreakand}{%
  \end{@IEEEauthorhalign}
  \hfill\mbox{}\par
  \mbox{}\hfill\begin{@IEEEauthorhalign}
}
\makeatother 
  
\usepackage{draftwatermark}
\SetWatermarkText{Preprint}   
\SetWatermarkScale{0.4}           
\SetWatermarkColor[gray]{0.9}     
\SetWatermarkAngle{45}            

\begin{document}

\title{Automated Generation of Microfluidic Netlists using Large Language Models

\thanks{
This material is based upon work supported by the National Science Foundation under Grant No. 2245494. Any opinions, findings, and conclusions or recommendations expressed in this material are those of the author(s) and do not necessarily reflect the views of the National Science Foundation.
}
}

\author{\IEEEauthorblockN{
Jasper Davidson\IEEEauthorrefmark{1}, 
Skylar Stockham\IEEEauthorrefmark{1}, 
Allen Boston\IEEEauthorrefmark{1}, 
Ashton Snelgrove\IEEEauthorrefmark{1}, \\ 
Valerio Tenace\IEEEauthorrefmark{2}, 
Pierre-Emmanuel Gaillardon\IEEEauthorrefmark{1}
} 

\IEEEauthorblockA{\IEEEauthorrefmark{1}Department of Electrical and Computer Engineering, University of Utah \\
\{jasper.davidson, skylar.stockham, allen.boston, ashton.snelgrove, pierre-emmanuel.gaillardon\}@utah.edu}

\IEEEauthorblockA{\IEEEauthorrefmark{2} Primis AI, Inc. \\
valerio@primis.ai}
}

\vspace{100mm}

\maketitle

\begin{abstract}

Microfluidic devices have emerged as powerful tools in various laboratory applications, but the complexity of their design limits accessibility for many practitioners. While progress has been made in microfluidic design automation (MFDA), a practical and intuitive solution is still needed to connect microfluidic practitioners with MFDA techniques. This work introduces the first practical application of large language models (LLMs) in this context, providing a preliminary demonstration. Building on prior research in hardware description language (HDL) code generation with LLMs, we propose an initial methodology to convert natural language microfluidic device specifications into system-level structural Verilog netlists. We demonstrate the feasibility of our approach by generating structural netlists for practical benchmarks representative of typical microfluidic designs with correct functional flow and an average syntactical accuracy of 88\%.

\end{abstract}


\IEEEpubidadjcol 
\section{Introduction}
\label{section:introduction}
Microfluidic devices (MFDs) are small-scale systems that precisely manipulate tiny volumes of fluids for chemical, biological, or physical processes. Recent advancements in MFDs have resulted in promising improvements to laboratory procedures across various scientific disciplines, leading to the development of high-performance ``lab-on-chip" devices \cite{convery_manufacturing_2019}. While these advances progress, the microfluidic design automation (MFDA) field is still in its early stages \cite{huang_review_2022}. MFDA tools demand both microfluidic expertise and additional proficiency in computer-aided design tools, algorithms, and manufacturing techniques. As a result, practitioners in microfluidic design have shown limited adoption of MFDA techniques \cite{mcdaniel_case_2017}. Meanwhile, Large Language Models (LLMs) have emerged as powerful tools, serving as a bridge between non-experts and high-productivity design across various domains.


In this work, we propose a methodology to leverage the capabilities of LLMs to generate microfluidic device designs from natural-language prompts provided by practitioners. The resulting output is a structural netlist composed of assembled predefined microfluidic primitives that achieve the functions outlined in the prompt. To the best of our knowledge, this represents the first application of LLMs for MFDA netlist generation, offering a preliminary and basic demonstration of its potential.


\begin{figure*}[!ht]
    \centering
    \includegraphics[width=1.0\textwidth]{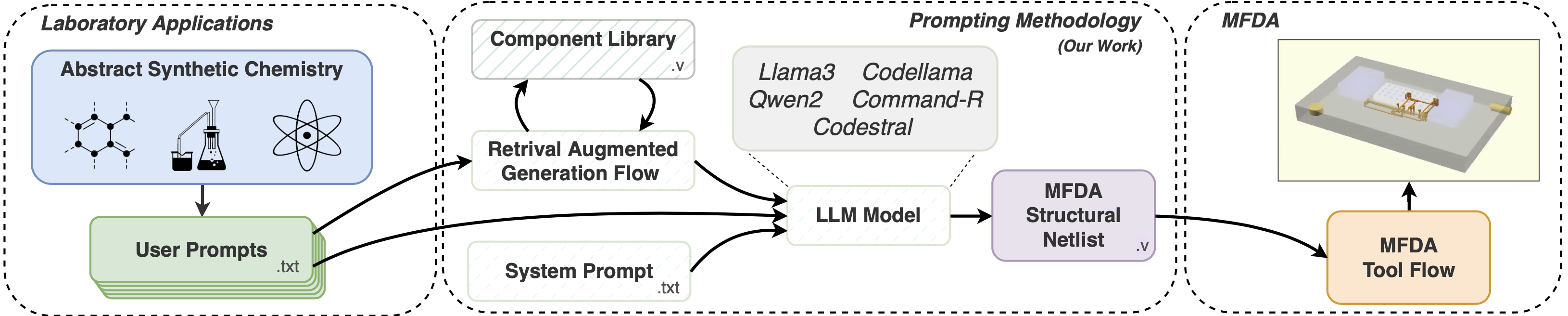}
    \caption{Microfluidic netlist generation methodology using LLMs, spanning from laboratory applications to Verilog netlists that are compatible with existing MFDA techniques.}
    \label{fig:flow}
\end{figure*}

We showcase the capabilities of our proposed introductorymethodology by creating structural netlists from benchmark prompts representative of standard microfluidic manipulations. Our results highlight that LLMs can accurately map the appropriate device primitives, establish correct connectivity between them, and adhere to the adopted netlist syntax. These findings mark an important first step in providing biologists and chemists with direct access to design tools for developing microfluidic devices. Using our proposed LLM prompting methodology, we produce microfluidic netlists with correct functional flow and average syntactical accuracy of 88\%. The proposed methodology is illustrated in \autoref{fig:flow}.


The remainder of this paper is organized as follows: Section \ref{section:background} reviews the state of the art in MFDA and LLM-based hardware design. Section \ref{section:methodology} presents the proposed prompting methodology. Section \ref{section:results} evaluates its performance, and Section \ref{section:conclusion} concludes the paper.

\section{Background and Related Work}
\label{section:background}
\subsection{Microfluidic Design Automation}
Traditionally, microfluidic designers have created devices using general-purpose computer-aided design (CAD) tools intended for mechanical engineering. These methods involve manually drawing channels and components as geometric shapes with dimensions, which are then used directly for manufacturing. This introduces two key challenges: a time-consuming, high-barrier design process for non-experts, and strong dependency on manufacturing limitations such as fabrication resolution, which constrains device and channel dimensions.

Microfluidic devices share many characteristics with digital electronics, and researchers have sought to apply EDA techniques to microfluidics \cite{huang_review_2022}. At a high level, both can be described as a netlist of internal components, their interconnections, and input/output ports. A desired but still missing capability is specifying devices using libraries of domain-specific components \cite{mcdaniel_case_2017}. As in EDA, providing manufacturing-dependent components introduces a crucial level of abstraction and portability.

Current state-of-the-art microfluidic design automation (MFDA) tools include Columba \cite{tseng_columba_2018}, Fluigi \cite{huang_fluigi_2015}, and OpenMFDA \cite{openmfda}. Columba and Fluigi use domain-specific languages to describe devices, while OpenMFDA leverages Verilog hardware description language (HDL) to tap into existing EDA tool flows. Nonetheless, all MFDA tools require designers to manually specify the syntax and low-level implementation details. While domain experts may be comfortable designing in structured textual formats, typical microfluidics practitioners are unlikely to adopt such approaches \cite{mcdaniel_case_2017}.

\subsection{LLMs for Hardware Development}
\label{section:related_works}
The use of LLMs for code generation is a growing, popular topic in current research. LLMs can potentially enhance engineers' productivity, including code completion \cite{liu_verilogeval_2023}, interactive design development \cite{blocklove_chipchat_2023}, and generating test benches for validation from natural language specifications \cite{aditi_specif_2023}.

This technique is also applied to hardware design, where early models were able to generate simple Verilog code snippets from human prompts \cite{pearce_dave_2020}. More recent works demonstrate a flow for optimizing LLMs to output Verilog HDL and define a standard for gauging the performance of a model, as well as proposals for improvement, achieving syntax-error-free code 65\% of the time \cite{thakur_llm_bench_2023}. CodeGen \cite{nijkamp2022codegen} highlights the challenges of synthesizing code from natural language input. 

While making user prompts more explicit can improve the output of LLMs, it may also increase prompt size and complexity, potentially hindering usability. Chip-Chat \cite{blocklove_chipchat_2023} describes an iterative process for hardware development via LLM rather than focusing on single prompt methodologies. To complement LLMs for code and HDL generation, recent works have established benchmark and evaluation data sets for LLMs in circuit hardware design  \cite{lu_rtllm_2023}\cite{liu_verilogeval_2023}.

These works highlight the utility of LLMs in the automated generation of hardware description languages within the EDA domain. However, at the time of writing, no published research has been found on the use of LLMs for generating microfluidic design specifications.

\section{Methodology}
\label{section:methodology}

\subsection{Describing Microfluidic Systems with Verilog}
\label{section:mtov}
We employ Verilog as a framework to represent components' structural composition and connectivity within the microfluidic device.
Primitive elements such as reservoirs, mixers, valves, and pumps are defined as cells, which contain a behavioral description of the device function with various metadata, such as geometric dimensions and flow rate, represented using Verilog parameters.
Predefined cells are interconnected in a structural netlist to construct a microfluidic system that executes the desired chemical manipulations.




\subsection{LLM Software Structure}
To develop our prompting methodology, we leverage existing open-source LLMs from Ollama \cite{OllamaWebsite}. Ollama provides a framework that consolidates various LLM models with diverse architectural characteristics, such as the number of parameters, layers, and context length, in one code base. Of the available models we select two models optimized for code generation (Codellama and Codestral) and two general purpose models (Llama3 and Qwen2). Additionally, we include the general-purpose Deepseek-R1 due to its state-of-the-art performance and efficiency in handling diverse workloads.

\begin{table}[h!]
    \centering
    \caption{Architectural Features of the Selected LLMs}
    \begin{tabular}{cccc}\toprule
         \multirow{2}*{Model} & \multirow{2}*{Parameters} & \multirow{2}*{Layers} & Context Length\\
         &&& (Number of Tokens) \\
         \midrule
         CodeLlama & 7 billion & 32 & 16,384 \\
         Codestral & 22 billion & 56 & 32,768 \\
         Llama3 & 70 billion & 32 & 8,192 \\
         Qwen2 & 72 billion & 80 & 32,768 \\
         DeepSeek-R1 & 32 billion & 60 & 131,072 \\\bottomrule
    \end{tabular}
    \label{tab:llms}
\end{table}

The number of parameters in a model reflects the weights in its layers, with more parameters enabling finer feature extraction. Network depth influences the ability to capture complex features, with deeper models improving performance but increasing overfitting risk. Context length determines how much of a prompt the model can process at once; prompts exceeding this length are split, causing each section to be considered separately. Tokenization defines the text units the model processes, with shorter context lengths reducing available contextual information. Table \ref{tab:llms} summarizes the key characteristics of the LLM architectures used in this paper.


\begin{table*}[h!]
    \centering
    \caption{Microfluidic domain benchmark prompts utilized to develop and refine our LLM prompting methodology.}
    \label{tab:Table2}
    \begin{tabular}{c|c|c}\toprule
         ~ & Benchmarks used for generating Prompting Methodology & Tests for: \\ \midrule
         1 & Take 2 solutions as input. Mix them together. & Basic interpretation of prompts \\
         2 & Take 5 solutions as input to the experiment module. Mix the 5 solutions together parallel. & Reasoning in parallel \\
         3 & Take 5 solutions as input to the experiment module. Mix the 5 solutions together sequentially. & Reasoning in sequence \\
         4 & Take 3 solutions as input. Dilute the first solution, then mix with the other two solutions. & Reasoning with dilution \\
         \multirow{2}{1mm}{5} & Put an algae solution into a heater. Mix the heated algae solution with itself to create a & \multirow{2}*{Abstract reasoning} \\
           & stirring effect. Pass through a membrane filter & \\ \bottomrule 

    \end{tabular}
\end{table*}

\begin{table*}[h!]
    \centering
    \caption{Microfluidic domain benchmark prompts applied to assess and validate our LLM prompting methodology.}
    \label{tab:Table3}
    \begin{tabular}{c|c|c}\toprule
         ~ & Test Benchmarks & Tests for: \\ \midrule
         6 & Take 2 solutions as input. Mix them together to create the output solution. & Basic interpretation of prompts \\
         7 & Take 4 solutions as input. Mix the 4 solutions together sequentially to create the output. & Reasoning in sequence \\
         8 & Take 6 solutions as input. Mix the 6 solutions together in parallel to create the output. & Reasoning in parallel \\
         9 & Take two solutions as input. Dilute the first solution, then mix with the other solution. & Reasoning with dilution \\
         10 & Heat up a solution of water. Filter the water to purify it, then mix with a diluted solution of oil. & Abstract Reasoning \\ \bottomrule

    \end{tabular}
\end{table*}

\subsection{Prompt Methodology}

We propose a prompting methodology with two primary components: retrieval-augmented generation (RAG) and a system prompt. The user prompt, library context from RAG, and system prompt are combined and then given to the model as visualized in \autoref{fig:flow}. The RAG process compares the similarity of the user prompt to computes vector similarity to the library and retrieves the top-k relevant primitives as context. This process gives the model access to information outside its training through vector spaces, and bypasses the need to retrain a new model every time a module is added to the library. We created a library of Verilog modules describing microfluidic primitives for MFDA and used RAG to provide the model with relevant context about their functionality. At the same time, a system prompt is developed throughout this study to guide the model’s behavior for the specific target application.

Using the system prompt, we limit the model to generating structural Verilog code that includes only user-defined primitives from the RAG library. This ensures the tool does not produce behavioral Verilog or functionality outside the microfluidic domain and prompts the user if certain requests are restricted based on the available set of primitives. In addition to giving the AI model further insight into the workings of Verilog, the system prompt is able to direct the model towards certain lines of thinking (such as the most effective ways to utilize different classes of Microfluidic devices e.g. heaters, diluters, mixers, etc.) as well as encourage chain-of-thought reasoning. By stepping through each of the logical steps the model takes, it's easier for the model to identify logical inconsistencies in its own reasoning.

\begin{figure}[]
\centering
    \includegraphics[width=0.48\textwidth]{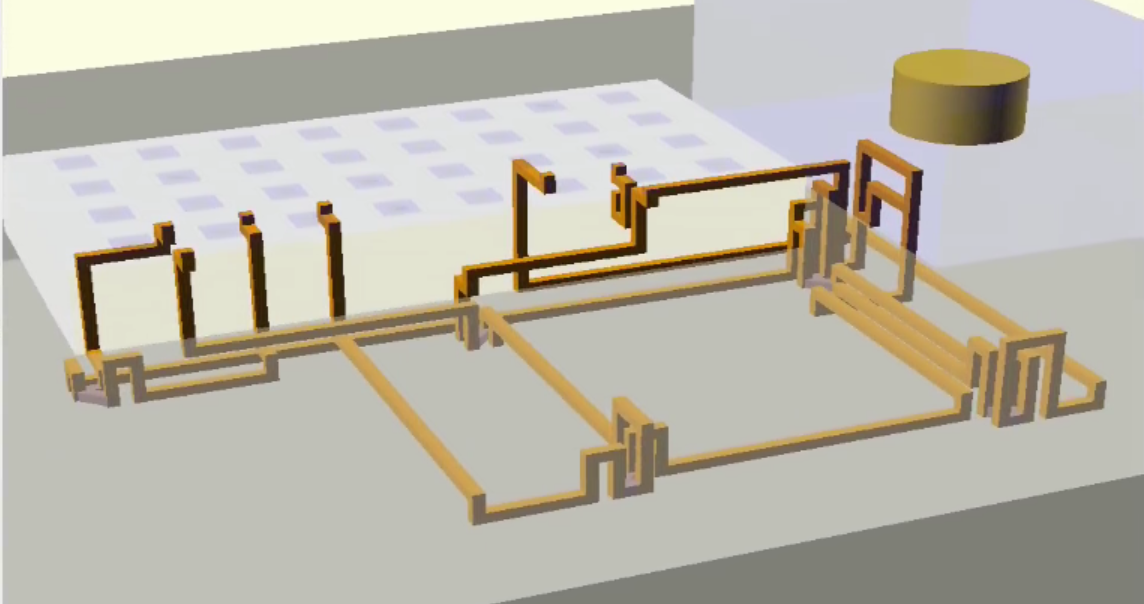}
    \caption{A three-dimensional rendering of a microfluidic chip designed to mix six solutions in parallel, produced using OpenMFDA with a Qwen2-generated system of six mixers.}
    \label{fig:synthesize}
\end{figure}


\subsection{Finishing the Workflow}

We integrate LLM netlist generation with the open-source MFDA toolchain, OpenMFDA, to demonstrate a complete workflow \cite{openmfda}.
OpenMFDA takes a structural Verilog netlist as input and generates a standardized microfluidic chip model ready for 3D printing.
The OpenMFDA rendering of a mixer tree is shown in Figure \ref{fig:synthesize}.
With this compatibility, we enable non-experts in the EDA or MFDA domains to generate ready-to-print devices using natural language as a frontend to an automated framework.

\section{Performance Evaluation and Analysis}
\label{section:results}

\subsection{Experimental Setup}
We assume that end users of this methodology lack the technical expertise to modify the model's output, making it crucial to generate a valid Verilog netlist on the first attempt. Furthermore, if this work were to extend beyond a single prompt, each subsequent interaction with the model would yield varying results, adding complexity and hindering our ability to report consistent quantitative outcomes effectively. Therefore, for our experiment, we selected five fundamental chemical manipulations commonly used in various laboratory procedures as user prompts, as detailed in \autoref{tab:Table2}. We then refined our prompt methodology to consistently generate representative, syntactically accurate structural netlists based on these foundational benchmarks. To assess the accuracy of the module generation, we utilized two key parameters. The ``microfluidic function" parameter assesses whether the output demonstrates a logically correct flow of components and accurately infers the appropriate microfluidic primitives according to fundamental microfluidic design principles. The ``Verilog syntax" parameter evaluates the output for accuracy and synthesizability as Verilog code.

\begin{table*}[ht!]
    \centering
    \caption{Pass@1 (single-short success rate) on MFD benchmarks testing for Verilog syntax and microfluidic functionality.}
    \label{tab:Table4}
    \begin{tabular}{c|ccccc|cccccc}\toprule
        Specification& \multicolumn{5}{c|}{Verilog Syntax} &\multicolumn{5}{c}{Microfluidic Function}\\ \midrule
        Pass@1[\%]   & CodeLlama & Codestral & Llama3 & Qwen2  & DeepDeek-R1   & CodeLlama & Codestral & Llama3 & Qwen2  & DeepDeek-R1 \\ \midrule
        Benchmark 6  &  80\%     & 100\%     & 100\%  & 100\%  & 100\%         &  80\%     & 100\%     & 100\%  & 100\%  & 100\% \\
        Benchmark 7  &  80\%     & 80\%      & 100\%  & 100\%  & 100\%         & 100\%     & 100\%     & 100\%  & 100\%  & 100\% \\
        Benchmark 8  &  0\%      & 20\%      & 60\%   & 80\%   & 100\%         &   0\%     &  40\%     &  80\%  & 100\%  & 100\% \\
        Benchmark 9  &  80\%     & 100\%     & 60\%   & 100\%  & 80\%          & 100\%     & 100\%     & 100\%  & 100\%  & 100\% \\
        Benchmark 10 &  0\%      & 100\%     & 60\%   & 60\%   & 60\%          &   0\%     & 100\%     &  80\%  & 100\%  & 100\% \\ \bottomrule
    \end{tabular}
\end{table*}

We identified the following as the most common points of error and subsequently structured our system prompt to specifically address these issues:

\begin{enumerate}
    \item Complete Verilog Initialization: ensuring adherence to proper Verilog syntax, ultimately determining if the netlist is compatible with existing tools that accept Verilog.
    \item Correctly Defined Primitives: ensuring the LLM property infers the correct microfluidic primitives, given the steps listed in the chemical manipulations.
    \item Correct Connections: validating whether the LLM accurately connects the ports of each primitive to implement the final microfluidic system.
    \item Correct Component Flow: determining whether the resulting netlist accurately emulates the specified benchmark.
\end{enumerate}


\subsection{Experimental Results and Discussion}

\subsubsection{Prompting Methodology Outcomes}
To assess the developed prompting methodology, we evaluate a new set of test benchmarks (user prompts) detailed in \autoref{tab:Table3}. Each benchmark was executed five times, and the generated netlists were evaluated based on the specified Verilog syntax and microfluidic function parameters. These results are compiled in \autoref{tab:Table4}, presented as a percentage of passing outputs. We observed that, in some cases, the correct microfluidic primitives were inferred, but the Verilog syntax was incorrect. Conversely, there were instances where the generated netlist was syntactically correct yet did not conform to the defined microfluidic function. 


On average, the microfluidic correctness of each model's output was better than their syntactical correctness. This uncovers a limitation of the prompting methodology, which struggled to enable the LLM to apply correct syntax across similar benchmarks. The system prompt included sections aimed at teaching the LLM proper syntax, such as example prompts followed by correct netlists and specifications for calling modules and creating wires. One way to address these syntactical errors is to fine-tune the model on the specifics of Verilog syntax. Alternatively, including a more comprehensive library of Verilog syntax in the RAG procedure could also help improve output quality without further extending the system prompt.

\subsubsection{Comparison between LLMs}
Qwen2 and DeepSeek-R1 outperformed all other models, benefiting from the highest number of parameters and the largest context size. However, Codestral, despite having fewer parameters than Llama3 and CodeLlama, performed better due to its context length being four times larger. This is crucial given the extensive system prompt, as a larger context size allows the LLM to connect different parts of the prompt and understand their relationships.


In Table \ref{tab:Table4}, our most important metric, correct component flow, passes for all but one benchmark. This shows that large-language models that are \textit{not} fine-tuned are able to perform extremely well in the design of microfluidic architecture. While this paper focuses on using currently open and available LLMs, building a model fine-tuned for Verilog syntax and microfluidic design automation could see syntactical accuracy improved beyond the 88\% demonstrated by Qwen2 and Deepseek-R1.

\vspace{-1mm}

\subsection{Future Work}
Our results demonstrate that the LLMs are capable of creating basic systems with user prompts; future work can be focused on creating more robust models for syntactical issues which will allow for more complex systems to be built with less detailed user prompts.

Recent works have proposed effective automatic verification techniques to produce near-perfect syntax for HDL \cite{islam2024edaawarertlgenerationlarge}. These works use a feedback system where each output is passed through an EDA tool, which checks it for syntactical errors. The prompt is automatically iterated with corrective prompts. This process is repeated until the final output is syntactically correct. Implementing a similar verification system could significantly increase the syntactical accuracy of models. Combined with the higher functionality of our RAG system which is tuned to allow greater functional accuracy, LLMs could become extremely accurate and consistent for MFDA netlist generation.

\section{Conclusion}
\label{section:conclusion}


We have presented the initial and foundational demonstration for generating system-level descriptions of microfluidic devices by using a well-adopted hardware description language and advanced ``language-to-hardware" capabilities of existing open-source LLMs. Our LLM prompting methodology achieved netlists with combined microfluidic functionality and Verilog syntactical accuracy of 88\%. This approach offers a streamlined and accessible method for researchers across various fields to harness the power of microfluidic platforms without requiring specialized knowledge of MFDA tools and corresponding device descriptions.

\bibliographystyle{IEEEtran}
\bibliography{references/IEEEabrv.bib,references/refs.bib}

\end{document}